# Speech-Based Cognitive Screening: A Systematic Evaluation of LLM Adaptation Strategies

**Fatemeh Taherinezhad*, MS[1]; Mohamad Javad Momeni Nezhad*, MS[1]; Sepehr Karimi, MS[1]; Sina Rashidi, MS[1]; Ali Zolnour, MS[1]; Maryam Dadkhah, MS[1]; Yasaman Haghbin, MS[1]; Hossein AzadMaleki, MS[1]; Maryam Zolnoori, PhD[1,2]**

[1]Columbia University Irving Medical Center, New York, NY

[2]School of Nursing, Columbia University, New York, NY, 10027 United States

*Authors contributed equally

**Corresponding Author:**
Maryam Zolnoori, PhD
Columbia University Medical Center,
School of Nursing Columbia University
560 W 168th St, New York, NY 10032
E-mail: mz2825@cumc.columbia.edu; m.zolnoori@gmail.com

**Number of Words:** 4700

**Number of Tables:** 1 (4 including appendix)

**Number of Figures:** 7

# Abstract

**Background:** Over half of U.S. adults with Alzheimer's disease and related dementias (ADRD) remain undiagnosed. Speech-based screening algorithms offer a scalable approach, but the relative value of large language model (LLM) adaptation strategies is unclear.

**Objective:** To compare LLM adaptation strategies for ADRD detection from the DementiaBank speech corpus using both text-only and multimodal models.

**Methods:** We analyzed audio-recorded speech from 237 participants and report performance on a held-out test set (n=71). Nine text-only LLMs (3B–405B; open-weight and commercial) and three multimodal audio–text models were evaluated. Adaptations included: (i) in-context learning (ICL) with four demonstration selection policies (most-similar, least-similar, class-centroid/prototype, random); (ii) reasoning-augmented prompting (self-/teacher-generated rationales, self-consistency, Tree-of-Thought with domain experts); (iii) parameter-efficient fine-tuning (token-level vs. added classification head); and (iv) multimodal audio–text integration. The primary outcome was F1 for the cognitively impaired (CI) class; AUC-ROC was reported when available.

**Results:** Class-centroid (prototype) demonstrations achieved the highest ICL performance across model sizes (F1 up to 0.81). Reasoning primarily benefited smaller models: teacher-generated rationales increased LLaMA-8B from F1 0.72 to 0.76; expert-role Tree-of-Thought improved its zero-shot score from 0.65 to 0.71. Token-level fine-tuning produced the highest scores (LLaMA 3B: F1=0.83, AUC=0.91; LLaMA 70B: F1=0.83, AUC=0.86; GPT-4o: F1=0.80, AUC=0.87). A classification head markedly improved MedAlpaca 7B (F1 0.06 to 0.82), indicating model-dependent benefits of this approach. Among multimodal models, fine-tuned Phi-4 Multimodal reached F1=0.80 (CI) and 0.75 (cognitively normal) but did not exceed the top text-only systems.

**Conclusions:** Detection accuracy is influenced by demonstration selection, reasoning design, and tuning method. Token-level fine-tuning is generally most effective, while a classification head benefits models that perform poorly under token-based supervision. Properly adapted open-weight models can match or exceed commercial LLMs, supporting their use in scalable speech-based ADRD screening. Current multimodal models may require improved audio–text alignment and/or larger training corpora.

**Keywords:** Cognitive impairment detection; Speech-based screening; Large language models adaptation; In-context learning; Reasoning-augmented prompting; Fine tuning; Multimodal speech-text analysis;

# INTRODUCTION

Alzheimer's disease and related dementias (ADRD) pose a significant public health challenge, currently affecting approximately five million individuals, or 11% of older adults in the United States[1,2,3]. This number is projected to rise to 13.2 million by 2050[4], underscoring the need for early, scalable detection strategies. Despite national efforts, over half of individuals with ADRD remain undiagnosed and untreated[5–7]. To address this gap, the National Institute on Aging has prioritized the development of accurate, accessible screening tools[7,8].

A promising direction involves natural language processing (NLP) to analyze spontaneous speech, which may reveal subtle cognitive changes missed by conventional screening[9]. Picture description tasks, such as the "Cookie Theft" scene[10], are widely used to elicit language markers of early decline. Prior pipelines follow two main approaches: (i) engineering acoustic and linguistic features[11,12] (e.g., lexical diversity, syntactic complexity), and (ii) fine-tuning transformer encoders such as BERT[13] (for transcripts) and Wav2Vec 2.0[14] (for raw audio). While both strategies show promise, they require extensive feature engineering and large labeled corpora[15]—resources often lacking in clinical settings—limiting generalizability across dialects and institutions.[16]

Large language models (LLMs) offer new opportunities for cognitive impairment detection by modeling complex linguistic patterns, performing few-shot in-context learning[17], generating reasoning chains, and adapting via fine-tuning. LLMs show strong performance in clinical decision support (CDS) tasks[18–20], including detection of depression[21], anxiety[22], suicide risk[23], and medication-related errors[24]. Applications to cognitive impairment are emerging but remain limited—for example, using GPT-4 in zero-shot fluency scoring, GPT-3 embeddings for classification, or comparing GPT-3.5, GPT-4, and Bard[25] on DementiaBank[26] transcripts. These studies suggest feasibility but lack systematic comparisons of prompting methods, fine-tuning, and multimodal inputs.

We present the first comprehensive evaluation of state-of-the-art large language models (LLMs), including open-weight (LLaMA[27], Ministral[28], MedAlpaca[29], DeepSeek[30]) and commercial models (GPT-4o[31], Gemini 2.0 Flash[32]), for early detection of Alzheimer's disease and related dementias (ADRD) using the Pitt copus[33] of DementiaBank benchmark. Our study comprises four components: (1) in-context learning with demonstration selection to assess the impact of different sampling strategies; (2) reasoning-augmented prompting to evaluate whether structured reasoning enhances LLMs performance, particularly in smaller models; (3) parameter-efficient fine-tuning to improve classification accuracy beyond prompt-based methods; and (4) evaluation of multimodal LLMs that integrate audio and text to determine the added value of acoustic information.

# METHODS

**Figure 1** provides and overview of the methodology of the study.

## Dataset

This study analyzed audio recordings from the DementiaBank[26] picture-description task (**Table 1**). The corpus contains 237 participants—122 cognitively impaired (CI) and 115 cognitively normal (CN). Following the DementiaBank split, 166 participants (87 CI, 79 CN) formed the development set, and 71 (35 CI, 36 CN) constituted the held-out test set. All diagnoses were made by neurologists or certified cognitive specialists.

A validation set was drawn from the development data via stratified sampling on diagnosis, Mini-Mental State Examination (MMSE) score, gender, and audio duration, yielding 116 training and 50 validation subjects. Recordings were transcribed with Amazon Web Services (AWS) General Transcribe[34].

We denote the transcription of a subject $i$ in $S_i^C$, in which $S$ represents the subject and the superscript $C$ indicates the subject's cognitive status, with $C \in \{CI, CN\}$.

Participants were ≥ 53 years old; women comprised > 60 % of each group. MMSE scores ranged from 3–28 in CI (mild–severe impairment) and > 24 in CN. CN speakers produced more words on average, whereas CI speakers had longer recordings, suggesting slower speech or greater effort (**Table 1**).

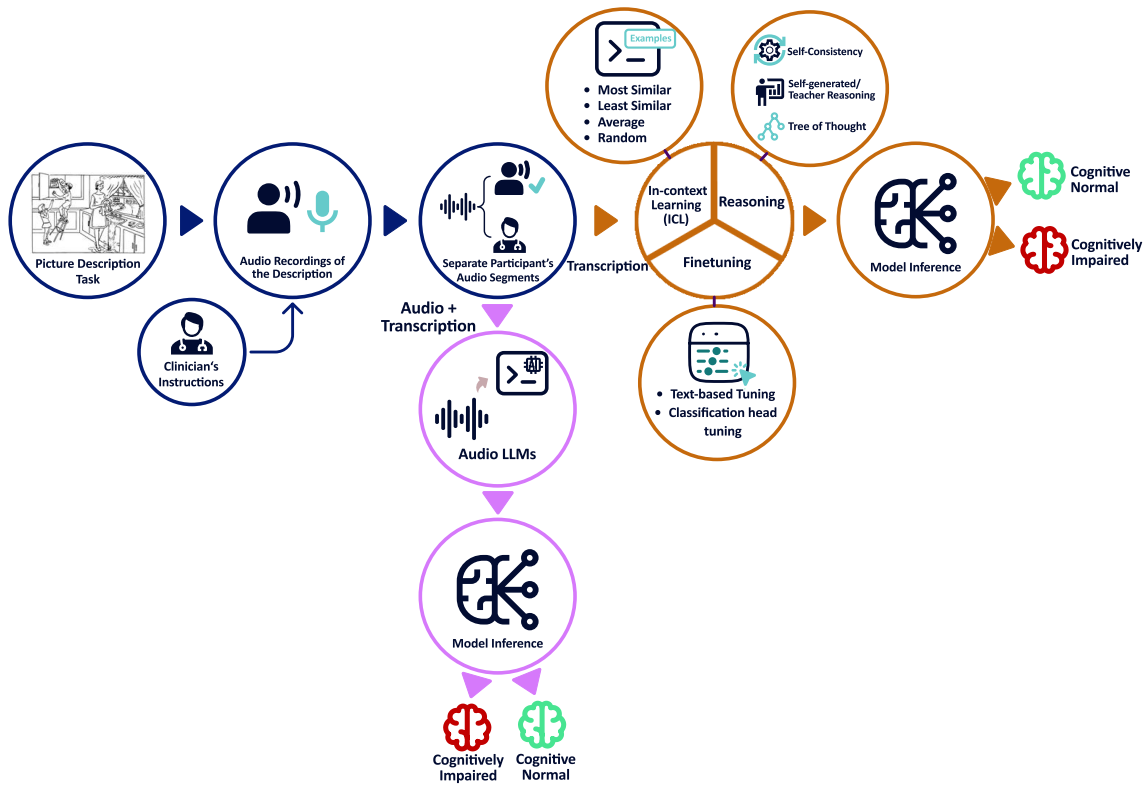

*Figure 1. Study workflow and evaluation framework for LLM-based ADRD detection. Participants complete the Cookie-Theft picture-description task, and their responses are audio-recorded under standardized clinician instructions. Recordings are segmented per speaker and transcribed with AWS. **Text-only (transcription) pipeline (orange)** includes: (1) In-context learning (ICL) with demonstration selection—few-shot examples are drawn from cognitively normal (CN) and cognitively impaired (CI) speakers using four sampling rules (most-similar, least-similar, group-average, random); (2) Reasoning-augmented prompting models receive self-generated/teacher rationales, self-consistency voting, or tree-of-thought chains; (3) Parameter-efficient fine-tuning using supervised text-based tuning and addition of a lightweight classification head. **Audio-enabled pipeline (purple)**: Raw speech and it transcript are fed to multimodal / audio LLMs, which directly encode acoustic and linguistic cues before inference, yielding the same binary outcome labels.*

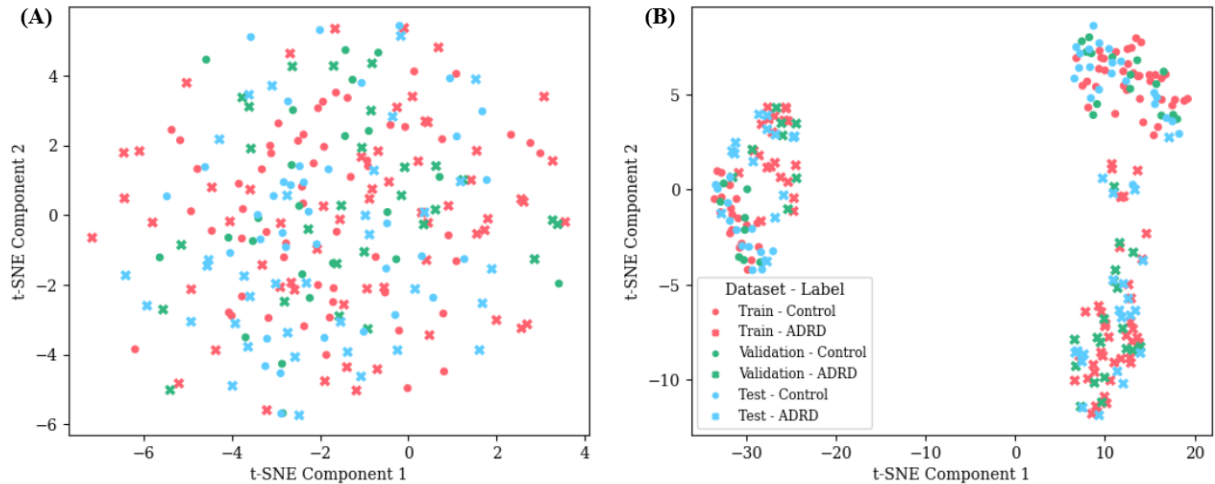

*Figure 2. t-SNE Visualization of Linguistic and Demographic Feature Spaces Across Dataset Splits. **A.** Two-dimensional t-SNE projection of word-level transcript embeddings. Points are color-coded by dataset split (train, validation, test) and diagnosis (control vs. ADRD). The extensive overlap indicates that all partitions occupy a comparable linguistic feature space, minimizing risk of distribution shift. **B.** t-SNE projection of participant-level metadata vectors combining age, MMSE, gender, recording duration, and word count. Three natural clusters reflect shared acoustic–demographic profiles, yet samples from every split and label are intermixed within each cluster, confirming balanced coverage of non-linguistic characteristics across partitions.*

To examine distributional similarity across partitions, we applied t-SNE to word-level embeddings from the transcripts (Error! Reference source not found..**A**) and to vectorized demographics (age, MMSE, gender, recording length, word count; (Error! Reference source not found..**B**), providing insight into overlap among training, validation, and test sets.

### Text-Only LLMs used in this study

We evaluated nine LLMs spanning diverse model sizes and training objectives. **GPT-4o** (text-only) served as a benchmark, representing a proprietary high-capacity model with advanced language understanding. **LLaMA 3.2 3B Instruct**[27] (LLaMA 3B), the smallest model, tested whether lightweight architectures can capture linguistic cues of cognitive impairment. **LLaMA 3.1 8B Instruct** (LLaMA 8B), a mid-sized model, was selected for its balance between efficiency and capacity to detect class-specific patterns. **MedAlpaca 7B**[29], fine-tuned on biomedical text, examined whether domain-specific pretraining enhances sensitivity to clinical language. **Ministral 8B**[28], optimized for efficient inference and strong text representation, evaluated performance of general-purpose mid-sized models. **LLaMA 3.3 70B Instruct** (LLaMA 70B) and **LLaMA 3.1 405B Instruct** (LLaMA 405B), large and ultra-large open-weight models, tested the impact of scale on capturing linguistic signals. **Gemini 2.0 Flash**[32], a commercial model optimized for low-latency inference and embedded reasoning, was included for its potential to detect cognitive impairment–related cues. **DeepSeek-R1**[30], trained on diverse multilingual data, assessed whether alternative training paradigms generalize across speaker populations.

### LLM Adaptation Strategies for Cognitive Impairment Detection

#### *Component 1: In-Context Learning with Demonstration Selection*

In-context leaning (ICL) prompts were composed of four elements: an Instruction ($IN$), a set of demonstrations ($DM$), the test input ($T_s$), and the corresponding output label ($L_s$). The model estimates the conditional probability:

$$P(L_s|T_s, IN, DM)$$

where each demonstration $DM_k = (Transcription_k, Label_k)$ and $Label_k \in \{CN = 0, CI = 1\}$

We began with a zero-shot baseline (i.e., $DM = \emptyset$), followed by few-shot experiments with $N = \{2, 4, 6, 8, 10, 12\}$ demonstrations. All prompts were standardized in structure and length across models to control for prompt-induced variance. The prompt presented in **Appendix 1**.

To examine how the type of demonstrations influences performance, we evaluated four selection strategies: (1) most similar, (2) least similar, (3) average similarity to class prototypes, and (4) random. Each strategy selected $N/2$ demonstrations from each class (CI and CN) to maintain balance.

*Table 1. Characteristics of the participants in "picture-description task"*

| **Attribute** | **Train** | | **Validation** | | **Test** | |
| --- | --- | --- | --- | --- | --- | --- |
| | Case | Control | Case | Control | Case | Control |
| **Participants** (N) | 60 | 56 | 27 | 23 | 35 | 36 |
| **Gender** (F/M) | 39/21 | 37/19 | 19/8 | 15/8 | 21/14 | 23/13 |
| **Age** (Mean ± Std) | 69.33 ± 7.14 | 66.27 ± 6.81 | 70.59 ± 6.01 | 65.48 ± 4.72 | 68.51 ± 7.12 | 66.11 ± 6.53 |
| Age Range | 53–79 | 54–80 | 60–80 | 56–74 | 56–79 | 56–78 |
| **Age Quartiles** (25%, 50%, 75%) | (65, 70, 75) | (60.75, 67, 71.25) | (65, 72, 76.5) | (63.5, 66, 68) | (63, 69, 74) | (61, 66, 70) |
| **MMSE** (Mean ± Std) | 17.80 ± 5.04 | 29.04 ± 1.13 | 16.63 ± 5.94 | 28.87 ± 1.22 | 18.86 ± 5.8 | 28.91 ± 1.25 |
| **MMSE Range** | 7–28 | 26–30 | 3–27 | 26–30 | 5–27 | 24–30 |
| **MMSE Quartiles** (25%, 50%, 75%) | (14.75, 18, 20) | (28, 29, 30) | (13.5, 17, 20.5) | (28.5, 29, 30) | (16, 20, 24) | (28, 29, 30) |
| **Recording Length** (Mean ± Std) | 87.20 ± 48.35 | 68.98 ± 25.85 | 88.52 ± 43.27 | 68.25 ± 25.43 | 79.42 ± 36.79 | 66.35 ± 28.17 |
| **Recording Length Range** | 35.26–268.49 | 22.79–168.61 | 39.91–219.5 | 26.16–121.47 | 28.39–150.15 | 22.35–135.68 |
| **Recording Length Quartiles** (25%, 50%, 75%) | (54.28, 75.93, 99.94) | (52.15, 67.6, 77.8) | (60.01, 80.24, 97.45) | (44.54, 67.77, 82.11) | (51.52, 70.20, 106.97) | (44.4, 66.04, 77.69) |
| **Word Count** (Mean ± Std) | 82.63 ± 43.32 | 114.43 ± 78.21 | 101.67 ± 55.49 | 111.39 ± 43.18 | 92.49 ± 57.38 | 111.72 ± 53.86 |
| **Word Count Range** | 22–189 | 21–523 | 31–284 | 54–197 | 27–256 | 45–243 |
| **Word Count Quartiles** (25%, 50%, 75%) | (51.25, 70.5, 106.25) | (67.25, 101, 139.75) | (67, 93, 118) | (78.5, 91, 147) | (50, 70, 120.5) | (63.5, 97, 168.25) |

Let $S_i^C$ denote the i-th transcript from class $C \in \{CI, CN\}$, and let $E(S_i^C) \in \mathbb{R}^d$ represent its semantic embedding computed using the BGE transformer embedding model[35]. For each test input $T_s$, we computed its embedding $E(T_s)$, and calculated cosine similarity with all candidate demonstrations (from the separated training dataset):

$$Score\ (S_i^C) = \cos\left(E_{ref}, E(S_i^C)\right)$$

Where the reference embedding $E_{ref}$ was defined differently for each strategy:

- Most Similar: $E_{ref} = E(T_s)$. Select the top $N/2$ samples per class with highest cosine similarity to the test input.
- Least Similar: $E_{ref} = E(T_s)$. Select the bottom $N/2$ samples per class with the lowest similarity to the test input.
- Average Similar: $E_{ref} = \bar{E}^C$, where $\bar{E}^C = \frac{1}{N^C}\sum_{k=1}^{N_c} E(S_k^C)$. Select the $N/2$ samples per class most similar to their class centroid, average of embeddings in each class.
- Random: Ignore similarity score and sample $N/2$ transcriptions per class uniformly at random.

Each strategy reflects a different hypothesis about which demonstrations best support generalization and reasoning:

- Most similar examples provide contextual alignment, enhancing sensitivity to subtle cues.
- Least similar examples increase linguistic variability, aiding generalization.
- Average similar samples serve as class prototypes, anchoring class distinctions.
- Random serves as a baseline for assessing the general value of demonstrations.

We computed F1-scores for the CI class on the validation set across all shot counts (N = 2–12). The optimal N for each strategy was selected on the validation set and used for final evaluation on the held-out test set.

### *Component 2 – Impact of Reasoning-Based Methods on Small LLMs*

To assess whether explicit reasoning enhances classification accuracy in cognitive impairment detection, we evaluated three reasoning-based prompting strategies across three resource-efficient LLMs: LLaMA 3B, LLaMA 8B, and Ministral 8B Instruct. To support these smaller models, we incorporated rationales generated either by the models themselves (self-generated) or by larger teacher models (GPT-4o and LLaMA 405B). The three strategies evaluated were:

#### (1) Reasoning-Augmented In-Context Learning (Reasoning-ICL)

Reasoning-ICL augments each demonstration with an explanatory rationale alongside the input transcription and label, enabling the model to better associate linguistic features with cognitive status. Rationales were sourced from: (1) self-generated explanations by the target model, and (2) teacher-generated rationales from a larger LLM (e.g., GPT-4o or LLaMA 405B). For each combination, we computed the F1-score for the cognitively impaired class on the validation set and selected the best-performing shot count. Final performance was then evaluated on the held-out test set using this optimal configuration. See **Appendix 2** for prompt design.

Formally, for each training transcription $S_k^C$, a rationale $Reason_k^x$ was generated, where $x \in \{self, teacher\}$ indicates the source of the explanation. Each demonstration is a triplet:

$$DM_k^{reason} = (S_k^C, Reason_k^x, Label_k) \text{ where } Label_k \in (CI, CN)$$

At inference, the target LLM received the test input $T_s$, a reasoning-specific instruction $IN_{reason}$, and a set of augmented demonstrations $DM_K^{reason}$ (chosen by the "Average" demonstration selection method introduced in component 1), then jointly generated both a rationale and classification label:

$$P(Reason_s, L_s | T_s, IN_{reason}, DM_K^{reason})$$

This framework enabled us to test whether adding structured rationales, generated by either the model itself or a more capable teacher model, improves the model's ability to detect cognitive impairment.

### (2) Self-Consistency with Teacher-Generated Reasoning

To assess whether reasoning-augmented in-context learning could be further improved by reducing variability in model outputs, we implemented the self-consistency[36] method, which aggregates predictions across multiple independently sampled inference runs using a fixed prompt. We restricted self-consistency to teacher-generated rationales, as results from *Results - Component2 – Self-Consistency with Teacher-Generated Reasoning* (see Results) showed that teacher-based reasoning using rationales from LLaMA 405B consistently outperformed self-generated rationales, GPT-4o rationales, and non-reasoning prompts across most shot counts.

For each shot count N = 2 to12, we used demonstrations augmented with teacher-generated rationales $DM_K^{reason}$. Each test input $T_s$ was processed five times using the same instruction ($IN_{reason}$) and the same set of demonstrations, under two temperature settings: 0.0 for deterministic decoding and 0.5 to introduce controlled randomness. Each run produced a pair: a generated rationale and a corresponding label ($Reason_s, L_s$)

$$P^i(Reason_s, L_s | T_s, IN_{reason}, DM_K^{reason}), i = 1, ..., 5$$

The final predicted label $\widehat{L_s}$ was computed by majority vote over the five predicted labels $L_s^{(1)}, ..., L_s^{(5)}$ using the below formula.

$$L_s = \underset{l \in (ADRD, Healthy)}{\operatorname{argmax}} \sum_{i=1}^{5} 1\{L_s^{(i)} = l\}$$

### (3) Tree-of-Thought (ToT) Reasoning

To evaluate a structured, multi-step reasoning approach beyond self-consistency, we implemented the Tree-of-Thought (ToT)[37] prompting framework. This method guides the model to break down decisions into intermediate steps, allowing it to generate and evaluate multiple reasoning paths before producing a final classification. By reasoning step-by-step, the model can retain, revise, or discard partial thoughts, potentially improving coherence and robustness.

We adopted a zero-shot setup to assess ToT's effectiveness independently of in-context demonstrations. For each test input, the model was prompted to reason from the perspective of three simulated experts, each generating a short sequence of reasoning steps. We tested two prompt formats:

(1) **Unspecified experts**: experts introduced with: "Imagine three different experts are analyzing a speech transcript…"

(2) **Domain-specific experts**: experts identified as a Language and Cognition Specialist, a Neurocognitive Researcher, and a Speech-Language Pathologist.

Each expert generated up to two sequential reasoning steps before providing a final classification. This corresponds to a tree with depth 2 and breadth 3. We capped the depth at two steps, as additional steps often led to repetitive or uninformative outputs.

This setup enabled evaluation of ToT as a standalone reasoning strategy without demonstrations, while maintaining consistent prompt structure and model size across methods. Full prompt templates are provided in **Appendix 3**.

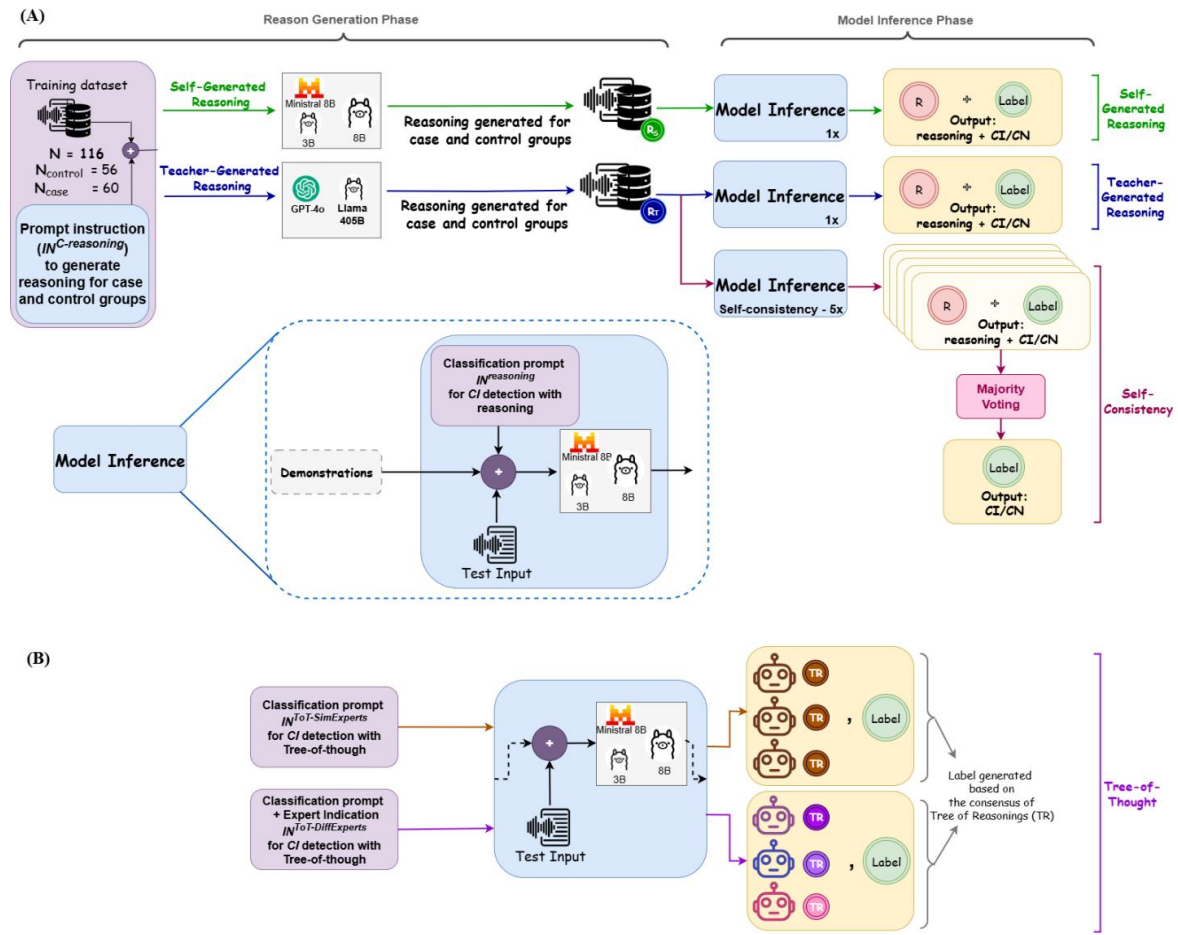

*Figure 3. Overview of reasoning pipeline. A. The methods include Self-generated Reasoning, Teacher-generated Reasoning, and Self-Consistency, where reasoning-augmented demonstrations are used for CI/CN classification and self-consistency aggregates multiple runs via majority voting. B. Tree-of-Thought pipeline, where the model is prompted to act as three experts, either unspecified or domain-specific, generate reasoning trees (TR) and determine the final label by consensus.*

### Component 3 – Fine-Tuning for Binary Classification

To assess whether task-specific adaptation improves model performance, we fine-tuned a subset of LLMs to classify transcripts as either cognitively impaired or cognitively normal. We implemented two approaches to fine-tuning.

#### (1) Token-Level Supervised Fine-Tuning

In this approach, classification was framed as a next-token prediction task. Each transcript was paired with a task-specific prompt (see **Appendix 4**), and the model was trained to generate the target label token, "AD" (CI) or "Healthy" (CN). The objective was token-level cross-entropy loss over the model's vocabulary, with the correct label token as the target.

Fine-tuning was applied to open-weight models (LLaMA 3B, LLaMA 8B, LLaMA 70B, MedAlpaca 7B, and Ministral) using Low-Rank Adaptation (LoRA)[38] for parameter-efficient optimization. We performed a grid search over LoRA rank (32, 64, 128), dropout (0.00, 0.05, 0.10), learning rate (2e-4), and batch size (4, 8, 16), with training epochs from 1 to 13. The best configuration for each model was selected based on F1-score for the cognitively impaired class on the validation set (**Appendix 5**).

For commercial models (GPT-4o, Gemini-2.0), we used API-level fine-tuning options, including learning rate multipliers, training epochs, and adapter or batch size where applicable. Hyperparameter choices were guided by API documentation and prior work. As with open-weight models, final settings were selected based on F1-score for the CI class on the validation set.

At inference, temperature was fixed at 0.0 for deterministic decoding. Where available, we also extracted class probabilities to compute threshold-independent metrics, specifically AUC-ROC. These probabilities were derived from SoftMax-normalized logits assigned to the "AD" and "Healthy" tokens in the output layer (see **Appendix 6** for more details).

#### (2) Classification Head Fine-Tuning

In this approach, we reframed the classification task by appending a lightweight classification head to the final hidden state of the LLM. The head consisted of three fully connected layers (output size: vocabulary dimension → 512 → 256 → 2), following the standard architecture used in Hugging Face implementations[39]. It was trained using binary cross-entropy loss to directly map hidden representations to class probabilities.

Unlike the token-level method, this approach decouples classification from language generation, allowing the model to learn class-specific features from its internal states rather than relying on token prediction. This method was applied only to open-weight models, where hidden representations are accessible.

Training inputs, prompts, and hyperparameter tuning followed the same procedures as in approach (1) (see **Appendix 4** and **Appendix 5**). During inference on the held-out test set, the classification head generated logits for each class, which were then converted into labels for evaluation.

### Component 4. Evaluating Multimodal LLMs as Classifier

To evaluate multimodal LLMs for cognitive impairment classification, we tested three state-of-the-art models using paired audio and transcripts. All models were prompted to process both modalities and output the patient's cognitive status. **Appendix 7** includes details of this prompt.

- **GPT-4o mini**[31]: OpenAI's closed-weight model supporting text and audio inputs. Due to limited access, we performed zero-shot inference using the API with temperature set to 0.
- **Qwen 2.5 Omni**[40]: Evaluated using two strategies:
    1. Zero-shot: Run with Hugging Face's recommended parameters (e.g., temperature = 1.0, top-k = 50, top-p = 1.0).
    2. Fine-tuning: Performed using LLaMA-Factory on training-set audio–transcript pairs. LoRA was used for efficient adaptation with recommended hyperparameters [40] (see **Appendix 8** for details).
- **Phi-4 Multimodal**[41]: Microsoft's multimodal successor to the Phi series.
    1. Zero-shot: Run with Hugging Face's recommended parameters (e.g., temperature = 1.0, top-k = 50, top-p = 1.0) using the same instruction prompt as Qwen and GPT-4o.
    2. Fine-tuning: Conducted via Hugging Face by a grid search over gradient accumulation steps, number of epochs and audio length and with recommended LoRA-based settings[41] (see **Appendix 8** for details).

For both Qwen and Phi-4, the number of epochs was selected based on validation F1-score for the cognitively impaired class.

### *Error analysis*

We selected fine-tuned GPT-4o and LLaMA 8B for error analysis because they showed consistent performance, and conducted two complementary analyses.

**Qualitative review.** We examined all misclassifications in the held-out test set—false positives (FP; cognitively normal predicted as impaired) and false negatives (FN; cognitively impaired predicted as normal)—by listening to the audio and reviewing AWS Transcribe transcripts. Each misclassified sample was labeled for: noisy audio (e.g., background noise, clipping) and missing/partial transcription.

**Quantitative analysis.** We computed 25 text-derived metrics across four domains—lexical richness (11), syntactic complexity (7), disfluency/repetition (2), and semantic coherence (5) (**Appendix 9**)—for all test samples. Distributions were compared using two-sided Mann–Whitney U test [42] for TP (true positive) vs. FN within the cognitively impaired group and TN (true negative) vs. FP within the cognitively normal group. A p-value < 0.10 was used to flag potential differences.

## Results

Throughout this section, F1-scores refer to the cognitively impaired class unless otherwise specified.

### Component 1: In-Context Learning with Demonstration Selection

**Figure 4.A** presents validation F1-scores for each LLM using 2–12 in-context demonstrations across four selection strategies. Demonstrations selected by *Average Similarity* to class centroids achieved the highest or joint-highest F1-scores in five models and ranked second in three others. The *Most Similar* strategy generally produced the next-best performance, with notable results for GPT-4o and Gemini-2.0. *Least Similar* examples yielded the lowest scores overall, except for MedAlpaca-7B and LLaMA-3B. *Random* selection showed minimal improvement over zero-shot,

suggesting limited benefit from unstructured examples. In larger models, performance gains plateaued after six demonstrations, indicating reduced sensitivity to demonstration quality, whereas smaller models remained more influenced by selection strategy.

**Figure 4.B** shows corresponding results on the test set. *Average Similarity* achieved the highest F1-scores in five models, including LLaMA 3B (0.73), Ministral 8B (0.73), LLaMA 70B (0.79), GPT-4o (0.81), and DeepSeek-R1 (0.79). *Most Similar* was optimal for LLaMA-8B (0.72), LLaMA-405B (0.80), and Gemini-2.0 (0.81). *Least Similar* continued to underperform, while MedAlpaca-7B again performed best with random samples (F1 = 0.67). These results highlight the importance of selecting representative, class-central demonstrations to enhance generalization in in-context learning.

**Component 2 – Impact of Reasoning-Based Methods on Small LLMs**

*(1) Reasoning-Augmented In-Context Learning (Reasoning-ICL)*

Validation results (Error! Reference source not found..**A**) indicated that adding rationales improved F1-scores across all three small LLMs compared with the no-reasoning baseline. Rationales generated by LLaMA 405B yielded the largest gains. With ten demonstrations, LLaMA 3B achieved an F1-score of 0.78 (vs. 0.64 baseline), while with twelve shots, Ministral 8B reached 0.77 (vs. 0.61) and LLaMA 8B reached 0.76 (vs. 0.72). Rationales from GPT-4o consistently outperformed self-generated rationales but were generally below LLaMA 405B across most shot counts.

Test-set results (Error! Reference source not found..**C**) were less aligned with validation trends. GPT-4o rationales produced the highest F1-score for Ministral 8B (0.72), while LLaMA 405B rationales yielded the best result for LLaMA 8B (0.78). Notably, LLaMA 3B performed best (0.66) with self-generated rationales. These discrepancies indicate that validation-set trends may not reliably reflect a model's generalization behavior, and that performance improvements from specific rationale sources or shot counts should be interpreted with caution.

*(2) Self-Consistency with Teacher-Generated Reasoning*

Using LLaMA 405B–generated rationales, we sampled multiple outputs per input and aggregated predictions by majority vote (Error! Reference source not found..**B**). Even with temperature = 0, repeated inferences produced slight variations, reflecting the inherent stochasticity of LLMs. Aggregating predictions via majority voting left LLaMA 8B performance unchanged, but reduced F1-scores from 0.78 to 0.75 for LLaMA 3B and from 0.77 to 0.74 for Ministral 8B. Using a moderate temperature (0.5) increased output variation without improving performance.

On the test set (Error! Reference source not found..**D**), though self-consistency improved model performance, it did not preserve the validation trend. Majority voting at T = 0.0 increased F1-score by 0.005–0.05 for LLaMA 3B, resulting 0.72, and LLaMA 8B, resulting 0.76, with respect to the performance of LLaMA 405 rationales, and lowered by 0.01 for Ministral 8B, reaching 0.67. Results at T = 0.5 were lower for the LLaMA family and only improved Ministral 8B's performance to 0.71. These findings highlight self-consistency as an effective strategy for mitigating prediction instability, an intrinsic property of LLMs, even under deterministic decoding settings.

## (1) Tree-of-Thought (ToT) Reasoning

In zero-shot classification, the three evaluated models—LLaMA 3B, LLaMA 8B, and Ministral 8B—achieved baseline F1-scores of 0.73, 0.55, and 0.57, respectively.

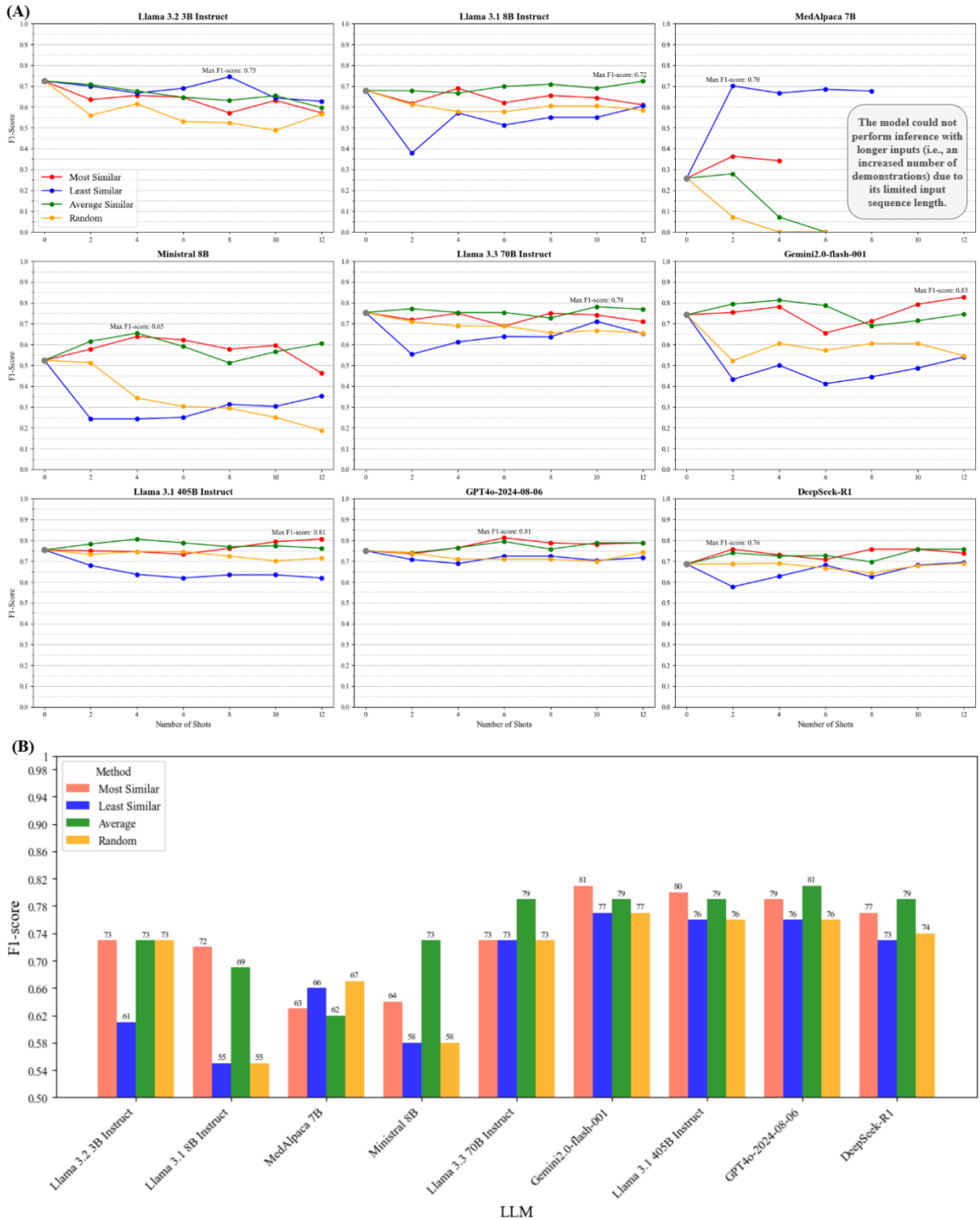

**Figure 4. Impact of demonstration selection strategies on in-context learning performance across large language models (LLMs). A.** Results for validation: F1-scores for 2–12 demonstrations show Average Similarity often outperforming other methods, with larger models plateauing after ~6 shots and smaller models showing greater sensitivity to selection quality. **B.** Results for test: Using optimal shot counts from (A), Average Similarity achieves the highest scores for most models, while Most Similar leads in a few cases. Numbers above bars indicate F1-scores ×100.

Applying Tree-of-Thought (ToT) prompting with unspecified expert roles altered performance to 0.59 (−0.10), 0.63 (+0.11), and 0.66 (+0.09) for LLaMA 3B, LLaMA 8B, and Ministral 8B respectively. When domain-relevant expert roles were incorporated, F1-scores increased to 0.68 (+0.09 vs. non-expert), 0.71 (+0.05), and 0.69 (+0.03), respectively. Compared with zero-shot, expert-role ToT produced notable gains for LLaMA 8B (+0.16) and Ministral 8B (+0.12), but remained below baseline for LLaMA 3B (−0.05). These findings indicate that expert-grounded prompting can enhance large model performance in cognitive impairment classification, whereas the smaller model, despite benefiting most from expert-role ToT relative to its non-expert counterpart, may lack the capacity for sustained multi-step reasoning.

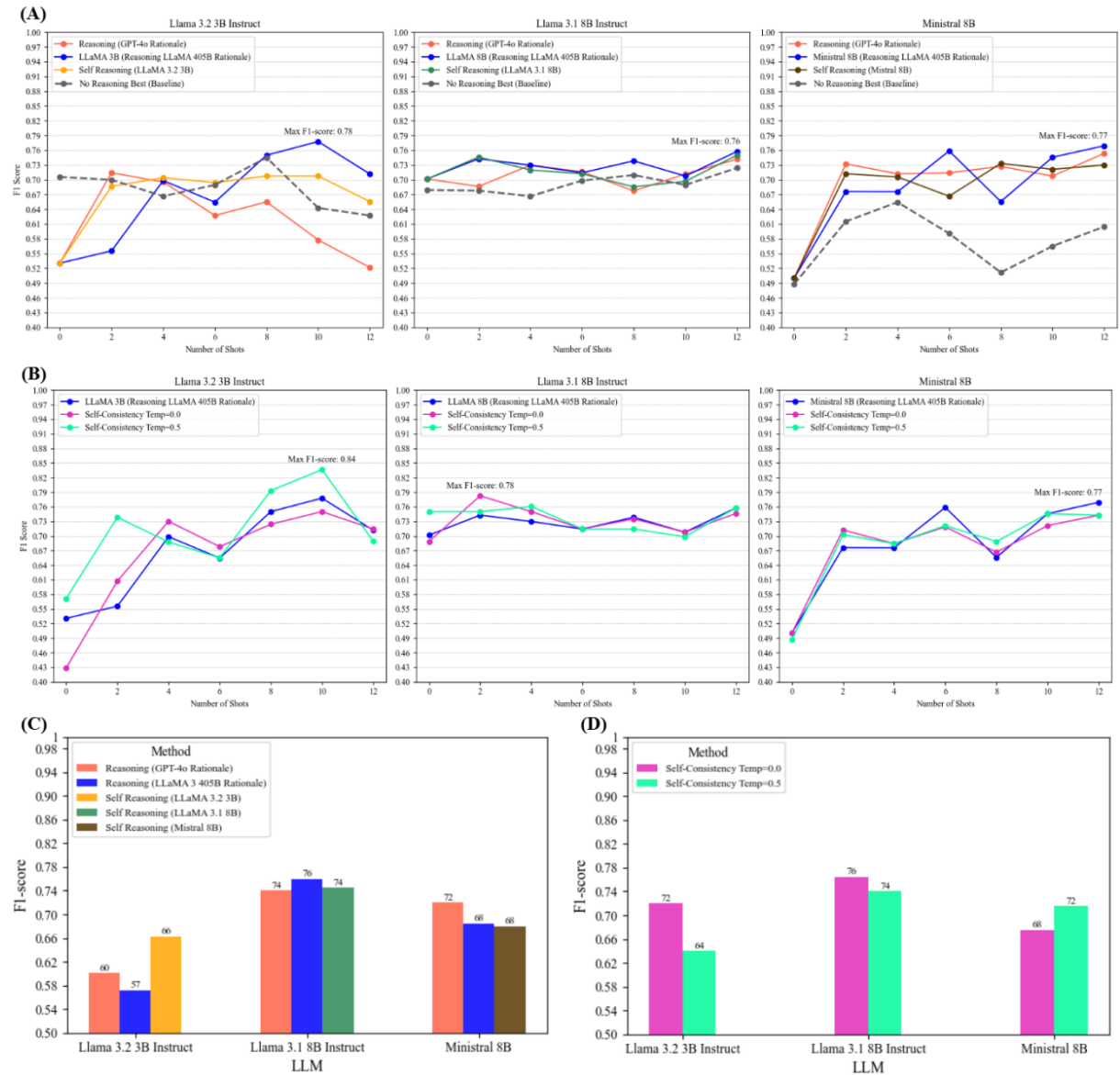

*Figure 5. Reasoning-augmented in-context learning and self-consistency performance on small LLMs. A.* Reasoning-Augmented ICL (Validation): *Adding rationales, especially those generated by LLaMA 405B, improved F1-scores over the no-reasoning baseline, with the largest gains in LLaMA 3B and Ministral 8B.* **B.** Self-Consistency (Validation): *Majority voting over multiple outputs with LLaMA 405B-generated rationales showed minor changes, with temperature adjustments having limited benefit.* **C.** Reasoning-Augmented ICL (Test): *Performance trends differed from validation; unlike validation, best scores varied by model and rationale source.* **D.** Self-Consistency (Test): *Majority voting slightly improved stability and accuracy for some LLaMA 3B and not for larger models. Numbers above bars indicate F1-scores ×100.*

## Component 3 – Fine-Tuning for Binary Classification

### *(1) Token-Level Supervised Fine-Tuning*

Error! Reference source not found. compares token-level supervised fine-tuning and classification-head fine-tuning across six models, reporting both AUC and F1-scores. Under token-level supervision (Error! Reference source not found..**A**), LLaMA 3B and LLaMA 8B achieved the highest AUCs (0.91 and 0.90) and corresponding F1-scores of 0.83 and 0.81, followed by GPT-4o (AUC = 0.87, F1 = 0.80), LLaMA 70B (AUC = 0.86, F1 = 0.83), Ministral 8B (AUC = 0.83, F1 = 0.79), and MedAlpaca 7B (AUC = 0.66, F1 = 0.06). Performance patterns indicate that smaller and mid-sized models achieved strong class separability, whereas MedAlpaca 7B underperformed, likely due to tokenization artifacts.

### *(2) Classification Head Fine-Tuning*

In contrast, classification-head fine-tuning (Error! Reference source not found..**B**) substantially improved MedAlpaca 7B (AUC = 0.92, F1 = 0.82, +0.76 improvement), while LLaMA 3B and LLaMA 70B declined to AUCs of 0.85 and 0.80 (F1 = 0.75 and 0.73, respectively), and Ministral 8B dropped to 0.71 (F1 = 0.66). LLaMA 8B also decreased to an AUC of 0.87 (F1 = 0.80). These results suggest that classification-head fine-tuning can markedly benefit models that perform poorly with token-level supervision, while models already performing well under token-level training may not gain, and can even lose, performance when switching to a classification-head approach.

## Component 4. Evaluating Multimodal LLMs as Classifier

- **GPT-4o Mini:** In the zero-shot setting, GPT-4o Mini achieved a high F1-score for cognitively impaired cases (0.70) but only 0.29 for cognitively normal cases, indicating substantial bias toward predicting impairment. Fine-tuning was not performed due to OpenAI's access limitations.
- **Qwen-2.5 Omni:** Zero-shot performance yielded an F1-score of 0.70 for cognitively normal cases and 0.54 for cognitively impaired cases, reflecting a reverse bias toward predicting cognitively normal. Fine-tuning did not improve performance and failed to address this imbalance.

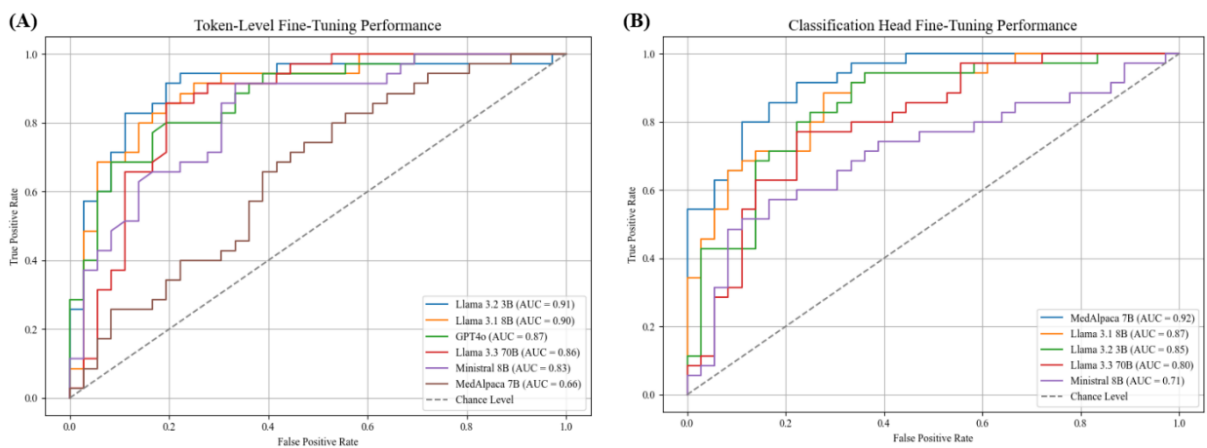

*Figure 6. Comparison of token-level and classification-head fine-tuning for binary classification on the test set. A. Token-level fine-tuning shows strong AUC performance for most models, with LLaMA 3B and LLaMA 8B leading and MedAlpaca 7B lagging. B. Classification-head fine-tuning markedly boosts MedAlpaca 7B but reduces performance for models already strong under token-level training.*

- **Phi-4 Multimodal:** Zero-shot performance was balanced, with F1-scores of 0.53 for cognitively impaired and 0.51 for cognitively normal cases. Fine-tuning led to substantial gains, reaching 0.80 for cognitively impaired and 0.75 for cognitively normal cases, the highest overall performance and largest improvement among all models.

These findings indicate that while GPT-4o Mini and Qwen-2.5 Omni performed reasonably in zero-shot mode, both exhibited strong class biases and limited benefit from fine-tuning. In contrast, Phi-4 Multimodal maintained balanced zero-shot performance and responded strongly to fine-tuning, underscoring the importance of task-specific training for robust cognitive health classification.

### Error analysis

**Misclassification overview.** On the held-out test set (n = 71), LLaMA-8B produced 6 false positives (TN = 30) and 7 false negatives (TP = 28); GPT-4o produced 8 false positives (TN = 28) and 7 false negatives (TP = 28). Three false positives and four false negatives overlapped across models. **Figure 7** summarizes error categories for LLaMA 8B and GPT-4o.

**Qualitative review.** Two problematic cases were excluded: one involved noisy audio with overlapping speech that produced an unrelated transcript and was misclassified by both models as an FP; the other had a missing transcription despite a high-quality audio file and was misclassified by GPT-4o as an FP. The remaining samples were used for quantitative analysis.

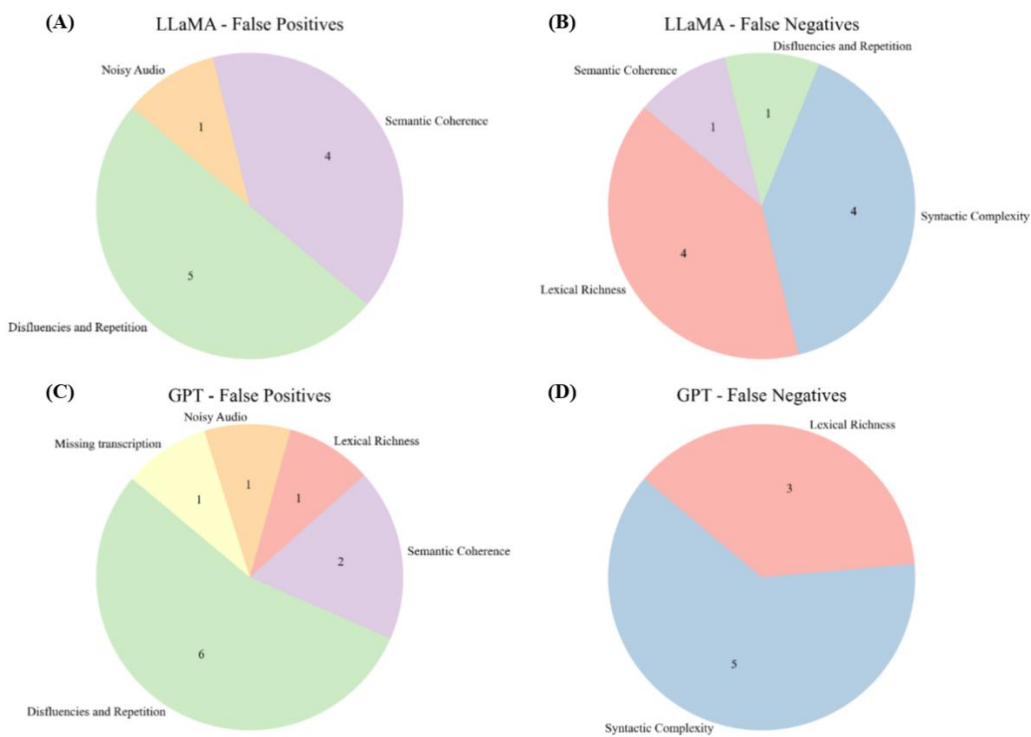

*Figure 7. Distribution of linguistic and technical issues contributing to model misclassifications on the test set. A. LLaMA 8B false positives were primarily due to disfluencies/repetition and semantic coherence. B. LLaMA 8B false negatives were mainly linked to lexical richness and syntactic complexity. C. GPT-4o false positives were dominated by disfluencies/repetition, with smaller contributions from semantic coherence and transcription-related issues. D. GPT-4o false negatives were largely associated with syntactic complexity and lexical richness.*

**Quantitative analysis.** Mann–Whitney U tests (p-value < 0.10) showed significant feature differences between correct and incorrect predictions. For GPT-4o, disfluencies/repetition differed for TP vs FN and TN vs FP, and semantic coherence differed for TN vs FP. For LLaMA 8B, syntactic complexity and semantic coherence differed for TP vs FN; lexical richness, semantic coherence, and syntactic complexity differed for TN vs FP. Together, these results suggest that misclassifications arise when a sample's linguistic profile resembles that of the opposite class.

A limitation of this analysis is that all ASR systems are prone to word insertions, repetitions, and truncations, which may have contributed to some of the observed errors and are reflected in our error categorization.

## Discussion

### Principal Results

This study presents the first comprehensive evaluation of multiple adaptation strategies, in context learning (ICL), reasoning-augmented ICLs, self-consistency, Tree-of-Though, and supervised fine-tuning across state-of-the-art open-weight and commercial large language models (LLMs) for detecting early cognitive impairment from speech transcripts (Pitt corpus of DementiaBank). Fine-tuning yielded the strongest performance: LLaMA 3B, LLaMA 70B, and LLaMA 8B achieved F1-scores of 0.83, 0.83 and 0.81, respectively, outperforming GPT-4o (F1 = 0.80). These results show that small open-weight models, when adapted to domain-specific tasks, can match or exceed commercial models, offering practical advantages in scalability and deployment.

In the context of in-context learning (ICL), demonstration selection strategy proved critical to performance. Demonstrations selected based on average similarity to class centroids, intended to reflect prototypical speech patterns of cognitively normal and impaired individuals, outperformed those based on most-similar, least-similar, or random selection. This effect was observed across both small and large models, with performance gains plateauing after six examples in larger models. These results highlight the importance of representative, class-central exemplars for guiding model generalization, especially in clinical tasks where linguistic variability may obscure diagnostic signals.

Teacher-generated rationales from LLaMA-405B or GPT-4o improved reasoning-augmented ICL for smaller models, increasing the F1-score of LLaMA 8B from 0.72 to 0.76. This suggests that teacher-generated reasoning can guide models toward better predictions, reducing adaptation costs by substituting for manually labeled examples. Self-consistency—aggregating predictions from repeated runs—boosted LLaMA 3B from 0.66 to 0.72 but offered limited benefit for larger models. These findings suggest that self-consistency mitigates prediction variability in smaller LLMs but is less impactful in models with more stable outputs.

Token-level fine-tuning outperformed classification-head adaptation for most models. An exception was MedAlpaca-7B, which performed poorly in the token-based setup (F1 = 0.06 for cognitive impairment class), likely due to its difficulty generating the correct label token during inference. However, when trained with a classification head, its performance improved substantially (F1 = 0.82 for cognitive impairment class). These results suggest that the effectiveness of fine-tuning strategies depends on model architecture and pretraining characteristics, particularly with respect to how reliably the model can produce discrete label tokens.

Multimodal LLMs underperformed relative to text-only models. In zero-shot settings, GPT-4o Mini and Qwen 2.5 Omni showed strong class bias, favoring cognitively impaired and cognitively normal classes, respectively. GPT-4o Mini could not be fine-tuned, and fine-tuning Qwen failed to improve performance, likely due to limited training data and poor audio-text alignment. Phi-4 Multimodal improved after fine-tuning (F1 = 0.80 impaired, 0.75 healthy) but still trailed top-performing text-based models. These findings suggest that current multimodal models require further adaptation for clinical speech classification.

## Comparison with Prior Work

The Pitt Corpus is a widely used benchmark for cognitive impairment detection from speech. Prior studies have used hand-crafted acoustic features (e.g., MFCCs), transformer-based embeddings (e.g., Wav2vec 2.0), rule-based linguistic metrics (e.g., LIWC), and BERT-based embeddings, achieving F1-scores between 0.70 and 0.87. Notable approaches include fine-tuning BERT-Large with ASR scores[43] (F1 = 0.85), combining multiple BERT variants with SVM[44] (F1 = 0.85), and ensembling logistic regression with fine-tuned BERT/ERNIE[45] (F1 = 0.82). More recent work has leveraged co-attention fusion[46] (F1 = 0.86), multimodal fusion with ChatGPT-derived embeddings[47] (F1 = 87.25), and cross-modal attention[48] (F1 = 0.84). In comparison, our LLM-based methods achieved F1 ≈ 0.81 with in-context learning and 80–83 with fine-tuning, demonstrating competitive performance with state-of-the-art systems.

These findings have important clinical implications. While biomarker-based tools (e.g., blood pTau217 and β-amyloid assays[49]) offer diagnostic value, they do not capture functional changes in everyday communication. Language impairment, an early sign of cognitive decline, remains poorly integrated into current screening workflows. LLM-based speech analysis offers a scalable, non-invasive approach to detect linguistic changes that complement biological markers. Integrating these tools into clinical settings[50] could enable earlier detection, improve decision-making, and broaden access to timely care.

Future work should combine LLM-based speech analysis with biological data, evaluate performance across diverse populations to ensure fairness, and address implementation challenges such as clinician acceptance, workflow integration, and regulatory compliance. LLMs, particularly when adapted with representative examples and reasoning strategies, offer a promising foundation for scalable cognitive screening.

## Limitations

This study has several limitations. First, the use of a single dataset (DementiaBank Pitt Corpus) limits generalizability to other speech corpora, languages, or dialects, potentially overlooking broader linguistic variability. Second, the limited training data, especially for multimodal models, may have restricted learning from acoustic inputs, making their underperformance difficult to interpret as model limitations alone. Third, reliance on automatic speech recognition (AWS) introduces transcription errors, particularly in impaired speech, which may disproportionately affect smaller models sensitive to input noise.

## Conclusion

This study provides the first systematic comparison of LLM-based adaptation strategies for detecting cognitive impairment from speech. Fine-tuned open-weight models matched or outperformed commercial LLMs and achieved performance comparable to advanced multimodal systems previously built on the benchmark Pitt Corpus. While current multimodal LLMs underperformed, results support LLM-based speech analysis as a scalable and effective approach for early cognitive screening.

## Acknowledgements

**Funding Source:**

R00AG076808 - "Development of a Screening Algorithm for Timely Identification of Patients with Mild Cognitive Impairment and Early Dementia in Home Healthcare" from National Institute on Aging."

**Author Contributions:**

- Fatemeh Taherinezhad: contributed to methodology design, data analysis, drafting the manuscript
- Mohamad Javad Momeni Nezhad: contributed to methodology design, data analysis, drafting the manuscript
- Sepehr Karimi: contributed to data analysis
- Sina Rashidi: contributed to data analysis
- Ali Zolnour: contributed to data analysis
- Maryam Dadkhah: contributed to figure design
- Yasaman Haghbin: contributed to methodology design
- Hossein AzadMaleki: contributed to methodology design
- Maryam Zolnoori: Leading conceptual model design and, drafting and critically revising the manuscript

### Data Availability:

The data is available in DementiaBank Dataset.

### Conflicts of Interest

None declared.

## Declarations

### Ethics Approval declaration

The data used in this study were obtained from the Pitt Corpus in the DementiaBank database, a publicly available resource hosted by TalkBank (https://dementia.talkbank.org/). The original data collection was approved by the Institutional Review Board of the University of Pittsburgh. As this study involved secondary analysis of de-identified data, no additional IRB approval was required.

### Human Ethics and Consent to Participate declaration

Not applicable.

## Abbreviations

CI: cognitive impairment

CN: cognitive normal

FP: false positive

FN: false negative

ICL: in-context learning

LLM: large language models

TP: true positive

TN: true negative

# Appendix

## Appendix 1: In-Context Learning with Demonstration Selection Prompt Design

To ensure consistency across all experiments in the few-shot setting, we employed a standardized prompt structure. The base prompt was as follows:

> *You are an expert in cognitive health and language analysis. You will analyze a spoken language transcript from a person describing the 'cookie theft' picture. This is not written text but a transcription of spontaneous speech. Analyze the provided transcript and classify it into one of two categories: 'Healthy' for a healthy cognitive state or 'AD' for Alzheimer's disease. Provide only the label ('Healthy' or 'AD') as the output. Do not include explanations or additional text. The output should be in JSON format, like {'label': 'predicted label'}.*

In the few-shot setting, demonstrations were introduced with the following prefix:

> *Here are some example cases for your guidance:*

This ensured a consistent prompt structure, with demonstrations presented before the test input to guide the model's prediction. Note that "Healthy" denotes cognitive normal and "AD" refers to cognitive impairment in the prompt.

## Appendix 2: Reasoning-Based Methods Prompt Design

The following prompt was used to generate reasoning for the training examples ($IN^{class-reasoning}$) using both the self-generated and teacher-generated methods, and these reasoned demonstrations were subsequently used for the in-context learning experiments.

> *You are an Alzheimer's Detection Assistant, a specialized model designed to identify potential indicators of Alzheimer's disease from spoken text transcripts. After analyzing each transcript and the corresponding labels provided, explain the rationale behind the categorization of each sample as either "Healthy" or "AD," indicating Alzheimer's disease. Keep the explanation short and concise. Format your explanation in JSON as follows:*
>
> *{"reason": "<Your explanation here>"}*

The following prompt is then used to inference the model and instruct it to generate reasoning. Just like the few-shot inference, the model is provided with examples when using demonstrations.

> *You are the Alzheimer's Detection Assistant. Your task is to analyze the linguistic features of the provided transcripts of spoken text for potential indicators of Alzheimer's disease. For each instance, label it as either "Healthy" or "AD" based on your analysis of the cognitive patterns in the speech, and first provide a **short reason** explaining why you classified it as Healthy or Alzheimer's disease. Start by briefly explaining your reasoning, then output your final decision in the following JSON format:*
>
> *{"reason": "provided reason", "label": "predicted label"}*

Note that "Healthy" denotes cognitive normal and "AD" refers to cognitive impairment in the prompt.

## Appendix 3: Tree-of-Thought Reasoning Prompt Design

The first prompt follows a zero-shot Tree-of-Thought structure in which the model simulates three unspecified experts reasoning collaboratively to classify a transcript as "Healthy" or "AD."

> Simulate three brilliant, logical experts collaboratively analyzing a transcript of spoken text. Each expert should explain their thought process in real-time, in a detailed and reflective manner, taking into account the prior explanations of the others and openly acknowledging any mistakes. At each step, each expert should refine and build upon the thoughts of the others whenever possible, and acknowledge their contributions. The discussion continues until a definitive answer is reached.
>
> Each expert's task is to analyze the linguistic features of the provided transcript for potential indicators of Alzheimer's disease. For each instance, the goal is to label the transcript as either "healthy" or "AD" based on their analysis of cognitive patterns in the speech.
>
> After all three experts have provided their analysis, review all three and provide either the consensus label or your best-guess label.
>
> The predicted label and analysis must be in JSON format, like {"analysis": "expert analysis", "consensus label": "predicted label"}

The second prompt extends this setup by assigning specific expert roles, each focused on a distinct aspect of language and cognition, to guide more structured and domain-informed reasoning.

> Imagine three different experts are analyzing a speech transcript to determine whether it reflects a "Healthy" cognitive state or indicates Alzheimer's disease ("AD"). Each expert is tasked with analyzing the linguistic features of the transcript for potential indicators of cognitive decline.
>
> All experts will write down one step of their reasoning, then share it with the group. They will proceed to a second step, if necessary. If any expert realizes their reasoning is flawed at any point, they will withdraw from the discussion.
>
> The experts are defined as follows:
>
> 1. **Language and Cognition Specialist**: Focuses on lexical diversity, syntactic complexity, and semantic coherence.
>
> 2. **Neurocognitive Researcher Studying Everyday Speech**: Examines pragmatic language use, discourse organization, and narrative structure.
>
> 3. **Specialized Speech-Language Pathologist**: Assesses speech fluency, articulation, and prosodic features.
>
> After all experts have completed their reasoning, the model must analyze their inputs and produce a final consensus classification.
>
> The response must be structured in JSON format as follows:
>
> {
>   "Language and Cognition Specialist": "...",
>   "Neurocognitive Researcher Studying Everyday Speech": "...",
>   "Specialized Speech-Language Pathologist": "...",
>   "Consensus Label": "..."
> }

Note that "Healthy" denotes cognitive normal and "AD" refers to cognitive impairment in the prompt.

## Appendix 4: Finetuning Prompt Design

The following prompt was used to fine-tune the models in both tuning approaches, text-generation and classification head, and was also used during inference to evaluate these fine-tuned models.

> You are an expert in cognitive health and language analysis. You will analyze a spoken language transcript from a person describing the 'cookie theft' picture. This is not written text but a transcription of spontaneous speech.
>
> Analyze the provided transcript and classify it into one of two categories: 'Healthy' for a healthy cognitive state or 'ADRD' for Alzheimer's disease and related dementias.
>
> Provide only the label ('Healthy' or 'ADRD') as the output. Do not include explanations or additional text.
>
> Text: {Transcript}
>
> Label:

Note that "Healthy" denotes cognitive normal and "AD" refers to cognitive impairment in the prompt.

## Appendix 5: Finetuning Details and Hyperparameters

This section provides the detailed settings used for finetuning the models on the Alzheimer's disease classification task using instruction-tuned text generation. The goal was to adapt each model to generate a binary label ("AD" or "Healthy") in response to transcript classification prompts.

For open-weight models, we fine-tuned using QLoRA, enabling efficient adaptation with minimal computational overhead. The following hyperparameters were tuned:

- Quantization: 4-bit quantization applied to reduce memory footprint.

- Rank & Alpha: Low-rank adapter dimensions (ranks of 16–128), with alpha scaling ($\alpha = 2 \times$ rank) to stabilize adapter outputs.

- Dropout: Regularization rates of 0, 0.05, and 0.10 explored to prevent overfitting.

- Learning Rate: Values of 2e-4 and 1e-4 tested, with 2e-4 yielding better convergence.

- Scheduler: Cosine annealing learning rate schedule to facilitate smooth convergence.

- Optimizer: PagedAdamW optimizer for efficient memory usage.

- Precision: Mixed-precision (FP16) training used across models. Batch Size & Gradient Accumulation: Adjusted based on model size to maintain effective batch sizes of 4–8.

- Epochs: Ranged from 1 to 13 depending on model size and validation performance.
- Max Sequence Length: Set to 1024 tokens to accommodate typical transcript lengths.

For API-based models (GPT-4o, Gemini-2.0), fine-tuning was limited to available parameters:

- Batch Size
- Number of Epochs
- Learning Rate Multiplier
- Adapter Size (for Gemini only)

The best performing parameters for each model are stated in the table below.

*Table 2. Selected hyperparameters for LLM finetuning*

| Model | QLoRA Rank | QLoRA Alpha | QLoRA Dropout | Batch Size | Grad Accum | Epochs | LR Multiplier | Adapter Size | Hardware |
|---|---|---|---|---|---|---|---|---|---|
| LLaMA 3B | 64 | 128 | 0.05 | 2 | 4 | 11 | - | - | NVIDIA A40 |
| LLaMA 8B | 64 | 128 | 0.1 | 4 | 2 | 12 | - | - | NVIDIA A40 |
| MedAlpaca 7B | 128 | 256 | 0.1 | 2 | 2 | 6 | - | - | NVIDIA A40 |
| Ministral 8B (2410) | 32 | 64 | 0 | 2 | 2 | 10 | - | - | NVIDIA A40 |
| LLaMA 70B | 16 | 32 | 0 | 1 | 4 | 9 | - | - | NVIDIA A100 |
| GPT-4o (2024-08-06) | - | - | - | 20 | - | 10 | 2.5 |  | API (OpenAI) |
| Gemini-2.0 | - | - | - | - | - | 20 | 5 | 8 | API (Google AI) |

## Appendix 6: Token Probability-Based Classification Details

For this method, models were fine-tuned using next-token prediction loss. During inference, the logits corresponding to the target tokens "AD" and "Healthy" were extracted from the model's output layer. The probabilities were computed using softmax:

$$P(Label) = \frac{e^{logit_{label}}}{e^{logit_{AD}} + e^{logit_{Healthy}}}$$

The final label was assigned as:

$$L_s = argmax(P(AD), P(Healthy))$$

For commercial models using the *top_logprobs* API, if either token was missing from the top 5, its probability was estimated as:

$$P_{missing} = 1 - \sum_{top\ 5} P_i$$

Where $P_i$ is the probability for token $i$ in the top five *logprobs* returned.

This ensured consistent label determination across models with limited output token visibility.

## Appendix 7: Multi-modal LLMs Fine-tuning Prompt

The following prompt yielded the best performance when applied to the fine-tuned models, Phi 4 Multimodal and Qwen 2.5 Omni. The same prompt was used during both training and inference.

> Transcription: "{transcription}"
>
> Based on the speech audio and its transcription, classify the speaker as dementia (Alzheimer's Disease or Related Dementia) or control (Cognitively Normal) with a single word: 'dementia' or 'control'.

Note that "control" denotes cognitive health and "dementia" refers to cognitive impairment in the prompt.

## Appendix 8: Multi-modal LLMs Fine-tuning Hyperparameters

The following hyperparameter values were used to fine-tune the two multimodal models.

*Table 3. Multi-modal LLMs tuning Hyperparameters*

| Model | LoRA Rank | Epochs | Batch Size | Gradient Accumulation Step | Learning Rate | Maximum Audio Length (seconds) | LR Scheduler | Warmup Ratio |
|---|---|---|---|---|---|---|---|---|
| Phi-4 | 320 | 3 | 1 | 32 | 2e-5 | 70 | — | — |
| Qwen-2.5-Omni | 64 | 5.0 | 1 | 4 | 1.0e-4 | — | cosine | 0.1 |

For LoRA adaptation and fine-tuning, we used all self-attention and MLP (feed-forward) blocks in Phi-4's language decoder and all transformer layers in Qwen-2.5-Omni, following the layer selections specified in the models' original papers[31,32].

## Appendix 9: Definitions of Linguistic Measures

This appendix defines the linguistic measures used to characterize Cookie Theft transcripts in our error analysis. Measures are grouped into Lexical Richness, Syntactic Complexity, Disfluencies and Repetition, and Semantic Coherence/Referential Clarity, and were computed from text to compare correctly (TP/TN) versus incorrectly (FP/FN) classified cases. Table below depicts a brief explanation of the features calculated.

*Table 2. Textual features calculated for quantitative error analysis.*

| Category | Feature | Description |
|---|---|---|
| **Lexical Richness** | Type–Token Ratio (TTR) | Proportion of unique words to total words, indicating vocabulary diversity. |
| | Root Type–Token Ratio (RTTR / Guiraud's) | Length-adjusted vocabulary diversity using unique words scaled by text length. |
| | Corrected Type–Token Ratio (CTTR / Carroll's) | Length-adjusted ratio of unique to total words to reduce text-length bias. |
| | Brunet's Index | Vocabulary diversity measure less sensitive to text length; lower values indicate richer vocabulary. |
| | Honoré's Statistic | Emphasizes rare words (hapax legomena) to capture lexical richness |
| | Measure of Textual Lexical Diversity | Assesses how consistently lexical diversity is maintained across the text. |
| | Hypergeometric Distribution Diversity | Probability-based estimate of lexical diversity accounting for sampling effects. |
| | Unique/Total Word Ratio | Fraction of unique words among all words, reflecting repetition vs. variety. |
| | Unique Word Count | Number of distinct words used in the text. |
| | Lexical Frequency | Average commonness of words based on a reference lexicon; higher values indicate more frequent words. |
| | Content Words Ratio | Proportion of nouns, verbs, adjectives, and adverbs, indicating information density. |
| **Syntactic Complexity** | Part-of-Speech Rate | Distribution of major POS categories, indexing grammatical variety. |
| | Relative Pronouns Rate | Proportion of relative pronouns (e.g., who, which, that), indicating use of subordinate relative clauses. |
| | Determiners Ratio | Proportion of determiners, reflecting specificity and clarity of reference. |
| | Verbs Ratio | Proportion of verbs, indexing predicate density and event description. |
| | Nouns Ratio | Proportion of nouns, indexing information density and entity mention. |
| | Negative Adverbs Rate | Proportion of negative adverbs, reflecting use of negation and more complex expression. |
| | Word Count | Total number of words as a proxy for elaboration and planning. |
| **Disfluencies and Repetition** | Speech Rate | Tempo of speech, indexing fluency. |
| | Consecutive Repeated Clauses Count | Number of back-to-back repeated phrases or clauses, indexing perseveration. |
| **Semantic Coherence** | Content Density | Amount of meaning-bearing content relative to text length. |
| | Reference Rate to Reality | Frequency of concrete references grounded in the picture or real-world entities/events. |
| | Pronouns Ratio | Proportion of pronouns, which can reduce clarity when antecedents are unclear. |
| | Definite Articles Ratio | Proportion of definite articles, indicating reference to specific, known entities. |
| | Indefinite Articles Ratio | Proportion of indefinite articles, indicating introduction of new or nonspecific entities. |